\documentclass[letterpaper, 10 pt, conference]{ieeeconf}

\IEEEoverridecommandlockouts

\usepackage{cite}
\usepackage{ulem}
\usepackage{xcolor}
\usepackage{gensymb}
\usepackage{acronym}
\usepackage{caption}
\usepackage{leftidx}
\usepackage{listings}
\usepackage{graphicx}
\usepackage{textcomp}
\usepackage{multirow}
\usepackage{booktabs}
\usepackage{subcaption}
\usepackage{algorithmic}
\usepackage{amsmath,amssymb,amsfonts}

\acrodef{VSLAM}{Visual SLAM}
\acrodef{STD}{Standard Deviation}
\acrodef{APE}{Absolute Pose Error}
\acrodef{SSE}{Sum of Squared Errors}
\acrodef{GCS}{Ground Control Station}
\acrodef{ROS}{Robot Operating System}
\acrodef{RGB-D}{Red Green Blue-Depth}
\acrodef{UAV}{Unmanned Aerial Vehicle}
\acrodef{RMSE}{Root Mean Square Error}
\acrodef{GPS}{Global Positioning System}
\acrodef{IMU}{Inertial Measurement Unit}
\acrodef{LiDAR}{Light Detection And Ranging}
\acrodef{SLAM}{Simultaneous Localization and Mapping}

\newcommand{\wrt}{w.r.t. }

\newcommand{\etc}{\textit{etc. }}
\newcommand{\eg}{\textit{e.g., }}
\newcommand{\ie}{\textit{i.e., }}

\newcommand{\etal}{\textit{et al. }}

\title{\LARGE \bf UAV-assisted Visual SLAM Generating Reconstructed 3D Scene Graphs in GPS-denied Environments}

\author{
    Ahmed Radwan$^{1}$, Ali Tourani$^{1}$, Hriday Bavle$^{1}$, Holger Voos$^{1}$, and Jose Luis Sanchez-Lopez$^{1}$
    \thanks{$^{1}$Authors are with the Automation and Robotics Research Group, Interdisciplinary Centre for Security, Reliability, and Trust (SnT), University of Luxembourg, Luxembourg. Holger Voos is also associated with the Faculty of Science, Technology, and Medicine, University of Luxembourg, Luxembourg. \tt{\small{ahmed.radwan.001@student.uni.lu}, \small{\{ali.tourani, hriday.bavle, holger.voos, joseluis.sanchezlopez\}}@uni.lu}}
    \thanks{*This work was partially funded by the Institute of Advanced Studies (IAS) of the University of Luxembourg (project TRANSCEND) and the Fonds National de la Recherche of Luxembourg (FNR) (project C22/IS/17387634/DEUS)}
    \thanks{*For the purpose of Open Access, and in fulfilling the obligations arising from the grant agreement, the author has applied a Creative Commons Attribution 4.0 International (CC BY 4.0) license to any Author Accepted Manuscript version arising from this submission.}
}

\begin{document}

\maketitle
\thispagestyle{empty}
\pagestyle{empty}

\begin{abstract}
Aerial robots play a vital role in various applications where the situational awareness of the robots concerning the environment is a fundamental demand.
As one such use case, drones in GPS-denied environments require equipping with different sensors (\eg vision sensors) that provide reliable sensing results while performing pose estimation and localization.
In this paper, reconstructing the maps of indoor environments alongside generating 3D scene graphs for a high-level representation using a camera mounted on a drone is targeted.
Accordingly, an aerial robot equipped with a companion computer and an RGB-D camera was built and employed to be appropriately integrated with a Visual Simultaneous Localization and Mapping (VSLAM) framework proposed by the authors.
To enhance the situational awareness of the robot while reconstructing maps, various structural elements, including doors and walls, were labeled with printed fiducial markers, and a dictionary of the topological relations among them was fed to the system.
The VSLAM system detects markers and reconstructs the map of the indoor areas, enriched with higher-level semantic entities, including corridors and rooms.
Another achievement is generating multi-layered vision-based situational graphs containing enhanced hierarchical representations of the indoor environment.
In this regard, integrating VSLAM into the employed drone is the primary target of this paper to provide an end-to-end robot application for GPS-denied environments.
To show the practicality of the system, various real-world condition experiments have been conducted in indoor scenarios with dissimilar structural layouts.
Evaluations show the proposed drone application can perform adequately \wrt the ground-truth data and its baseline.
\end{abstract}

\section{Introduction}
\label{sec_intro}

Employing \acp{UAV}, or simply drones, for a wide range of applications, such as monitoring, logistics, surveillance, remote sensing, data acquisition, and disaster management, has attracted the attention of researchers in recent decades \cite{mishra2020drone, dilshad2020applications, park2018forestry, krajewski2020round}.
These robots are equipped with various sensors, including \ac{GPS}, \ac{IMU}, and vision sensors, aiding them in navigation, accurate localization, path planning, obstacle avoidance, \etc \cite{vanhie2021indoor}
However, for indoor flights where the \ac{GPS} signals are missing, relying on other sensors, such as vision sensors (\ie cameras), is essential for the proper functioning of drones while performing pose estimation and localization \cite{akbari2021applications}.
Such sensors aid drones in performing various robotics tasks, including autonomous navigation and \ac{SLAM}, and improve the robots' Situational Awareness \cite{slamtosa}.

\begin{figure}[t]
    \centering
    \includegraphics[width=1.0\columnwidth]{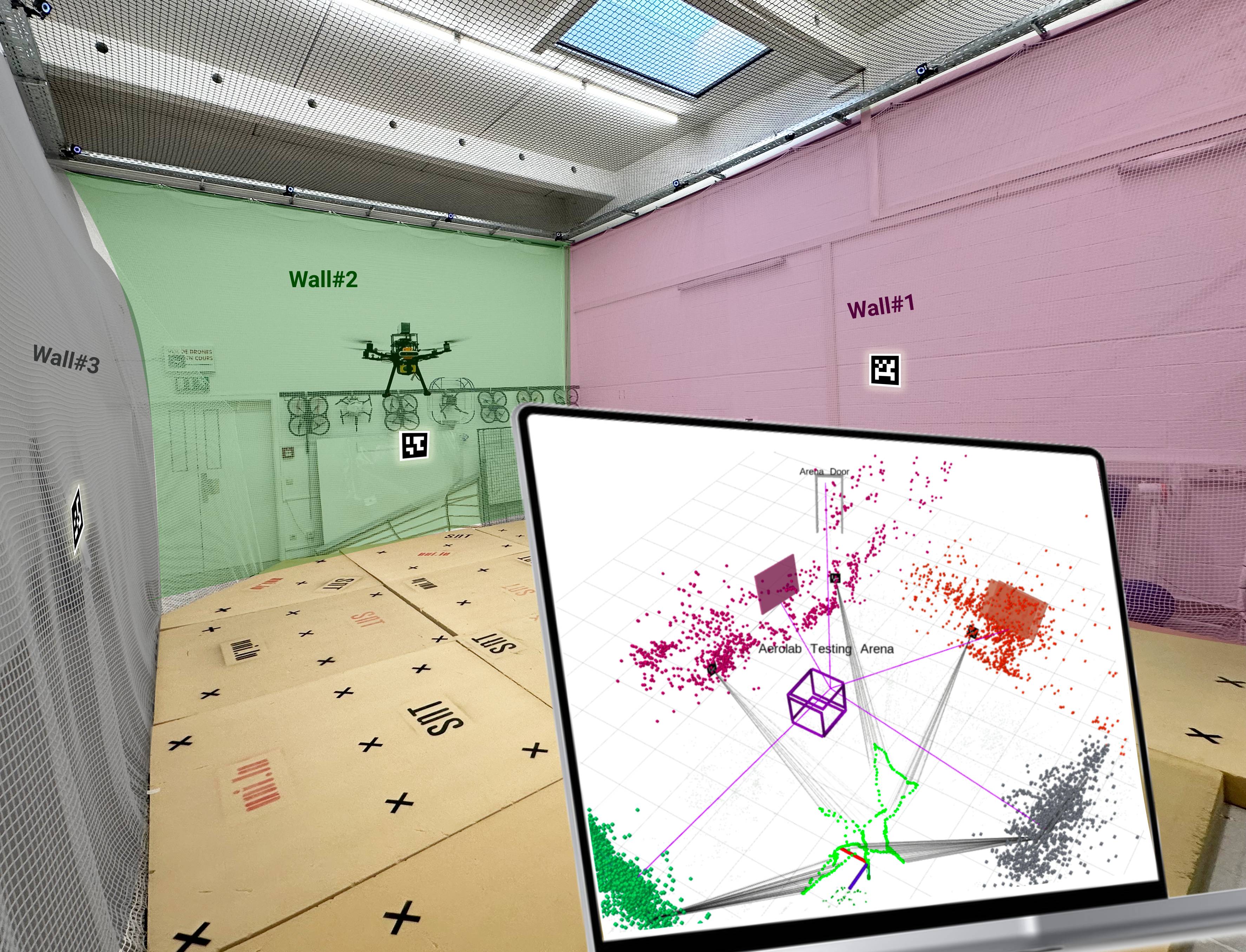}
    \caption{Reconstructing the map of an indoor environment, along with its 3D hierarchical representation and detected semantic entities generated by integrating a marker-based \acl{VSLAM} framework \cite{vsgraphs} on a drone.}
    \label{fig_overall}
\end{figure}

In this regard, some approaches generate meaningful 3D scene graphs of the environments \cite{3dsg, hydra} or incorporate \ac{SLAM} graphs with 3D scene graphs for richer generated maps representation \cite{sgraphs, sgraphsp} using \ac{LiDAR} sensors.
However, employing cameras for doing \ac{SLAM} and using the visual data provided by them for map reconstruction has resulted in the emergence of a new branch known as \acf{VSLAM} \cite{gao2021introduction}.
\ac{VSLAM} has become more trendy due to incorporating rich visual information for tracking, localization, and mapping steps, along with better scene understanding and semantic object recognition.
Particularly, when integrating semantic information and object-based topological relation is targeted, utilizing vision sensors is considered a more attractive approach compared to \acf{LiDAR}-based solutions \cite{vpsslam}.

When dealing with cameras as the primary sources of information for \ac{VSLAM}, there are some challenges in detecting and tracking visual features in the scene.
For instance, facing smooth and feature-less objects such as plaster walls and shiny ceramic tiled floors causes \ac{VSLAM} system to fail to track and lose localizing data.
One of the possible solutions to overcome the mentioned challenges is to employ fiducial markers as artificial landmarks placed on the environment to augment more data for robots \cite{kalaitzakis2021fiducial}.
We can see that various fiducial marker libraries, such as ArUco \cite{aruco} or AprilTag \cite{apriltag}, have been used in multiple marker-based \ac{VSLAM} methodologies, including UcoSLAM \cite{ucoslam}, sSLAM \cite{sslam}, and TagSLAM \cite{tagSLAM}.

This paper introduces an end-to-end aerial robotics application empowered by a marker-based \ac{VSLAM} framework (presented by the authors of the paper in \cite{vsgraphs}) and integrating it into a \ac{UAV} equipped with a mounted RGB-D camera.
The aim is to employ a drone flying in \ac{GPS}-denied environments to generate tightly coupled visual situational graphs of the areas, along with their low- and high-level semantic entities presented in 3D hierarchies.
Accordingly, the primary contributions of the paper in hand are summarized below:

\begin{itemize}
    \item Preparing a \ac{UAV} platform equipped with proper sensors and modules for semi-autonomous flight in \ac{GPS}-denied environments,
    \item Integration of a \ac{VSLAM} framework into the mentioned \ac{UAV} for real-time reconstruction of the maps of indoor areas, enriched with semantic and construction-level objects,
    \item Wrapping the \ac{UAV} and the \ac{VSLAM} framework into an end-to-end situationally aware robot,
    \item And performing real-world scenario experimental validations using the mentioned system.
\end{itemize}
\section{Related Works}
\label{sec_related}

This section contains state-of-the-art works related to the current research topics, covering the operation of \acp{UAV} for indoor environment map reconstruction and generating graph-based 3D representations.
Accordingly, studied research works are classified into three categories, listed below:

\subsection{Drone-Assisted Localization and Mapping}
Localization of drones and mapping the environments with their particular detected objects has many use cases targeted by drones.
In this regard, Raja \etal \cite{pfin} introduced a Particle Filter-based indoor navigation and mapping framework for \acp{UAV} that targets precision enhancement for localization and velocity estimation for collision avoidance.
In \cite{budiharto2021mapping}, authors introduced a \ac{SLAM} system for 3D models of environments using photogrammetric and situation mapping with Geographic Information Systems (GIS).
In another application, Bhatnagar \etal \cite{bhatnagar2020drone} introduced a deep learning-based approach for segmenting images captured by drones to map key vegetation communities.
Authors in \cite{sun2023indoor} introduced an indoor localization and tracking methodology based on Acoustic Inertial Measurement (AIM).
They use an improved Kalman filter algorithm to reduce location estimation errors and solve tracking challenges in \ac{GPS}-denied environments.
As another research, PILOT \cite{famili2022pilot} is a framework introduced for precise indoor localization in autonomous drones based on the relative geometrical analysis among transmitters and receivers of ultrasonic acoustic signals.
Similarly, ROLATIN \cite{famili2020rolatin} is another localization and tracking framework designed for indoor navigation using drones that rely on speaker-generated ultrasonic acoustic signals.

It can be seen that various sensor-based approaches to perform localization, tracking, and mapping exist, which highlights the versatility of the \acp{UAV} in addressing related issues.
Thus, with advancements in sensor technologies and sensor-fusion approaches, drones are considered potential solutions to be integrated with applications to create accurate maps.

\subsection{\ac{VSLAM} and 3D Scene Graphs}
3D scene graph techniques such as \cite{3d_scene_graph, 3D_dsg, scenegraphfusion} can generate high-level and optimizable hierarchical representations of environments and depict their recognized semantic entities with suitable relations among them.
However, unless other standalone frameworks, such as \textit{S-Graphs} \cite{sgraphs}, \textit{S-Graphs}+ \cite{sgraphsp} and Hydra \cite{hydra}, where the \ac{SLAM} maps and scene graphs are tightly coupled, many methodologies target only graph optimization.
Considering the introduced advantages of vision sensors, \ac{VSLAM} methodologies have evolved and become more mature and reliable in recent years.
To avoid repetition and save space in the current paper, the authors would like to refer to a survey conducted by them in \cite{tourani2022visual}, which provides insights into state-of-the-art approaches in \ac{VSLAM}, as well as its current trends and potential future directions.
Accordingly, reliable and versatile frameworks, such as various versions of ORB-SLAM \cite{orbslam, orbslam2, orbslam3}, have become the baseline for many \ac{SLAM} approaches that take advantage of vision sensors.
Although the geometric maps reconstructed by many methodologies are accurate, fetching semantic entities and adding them to the map can bring about more comprehensible maps and lead to improved optimization outcomes due to considering object-level data associations, as shown in works like \cite{vpsslam, blitzslam, qian2021semantic, doherty2019probabilistic}.

In this regard, a deep learning-based \ac{VSLAM} framework that fetches semantic information from the scene to perform multi-object tracking is presented in \cite{sun2022multi}, which augments semantic objects in the reconstructed map.
The primary challenge faced with the mentioned methodology is the potential for misclassification of the detected objects.
Guan \etal \cite{guan2020real} proposed a real-time semantic \ac{VSLAM} methodology, which adds point–object and object–object associations based on the scene segmentation output, in addition to the existing point–point association in ORB-SLAM2.
DS-SLAM \cite{yu2018ds} is another framework that utilizes SegNet for semantic information gathering and semantic mapping.
The mentioned approach can work in high-dynamic environments if appropriate processing hardware is provided.
Authors in \cite{yang2022visual} proposed another framework that emerged with a dynamic object removal module for generating static semantic maps.
Regardless of the ability of the mentioned methods to map various semantic elements in the environment, many of them may still face challenges due to misidentification and elements' pose estimation errors.
In this regard, adding structural and topological constraints among the detected semantic elements can lead to improved scene understanding, more accurate mapping, and further map optimization.

\begin{figure}[t]
    \centering
    \includegraphics[width=1.0\columnwidth]{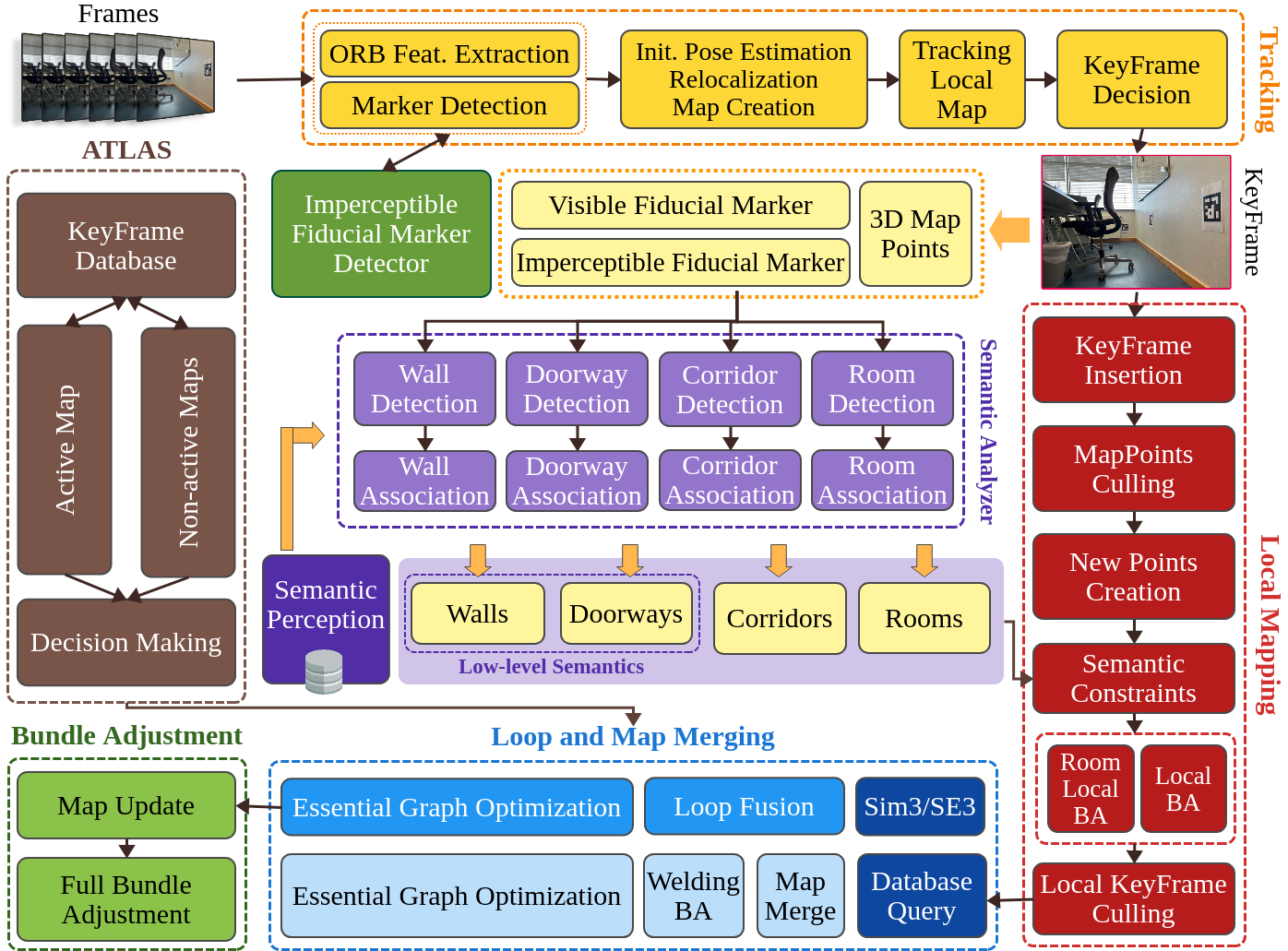}
    \caption{The primary system components and pipeline of the \ac{VSLAM} framework used in this research \cite{vsgraphs}.}
    \label{fig_pipeline}
\end{figure}

\subsection{Marker-based \ac{VSLAM}}
As fiducial markers are considered robust tools for augmenting semantic information, several marker-based \ac{VSLAM} frameworks have been developed for enhanced scene understanding and mapping.
Among this, UcoSLAM \cite{ucoslam} can perform \ac{VSLAM} tasks by employing the visual features obtained from both natural and artificial landmarks (\ie ArUco markers).
It contains a marker-based loop closure detector and supports keypoint-only, marker-only, and mixed modes.
sSLAM \cite{sslam} introduced by Romero-Ramirez \etal is another similar approach that employs customized markers for robust tracking.
The primary contribution of sSLAM is guaranteeing a substantial reduction in resource consumption and computation time while maintaining tracking accuracy.
TagSLAM \cite{tagSLAM} is another methodology that works in marker-only mode and requires having AprilTags in the scene for localization and tracking.

The identifiable gap in the mentioned approaches is reconstructing geometric maps with the aid of fiducial markers while ignoring their potential to augment semantic information.
Accordingly, a semantic version of UcoSLAM proposed in \cite{semuco} can simultaneously generate 3D scene graphs while performing \ac{SLAM}.
It can detect construction-level semantic entities using ArUco markers and utilize their topological relations for optimization.
In a similar work, the authors of this paper introduced another marker-based semantic \ac{VSLAM} framework \cite{vsgraphs} that provides more accurate reconstructed maps and generates three-layered optimizable hierarchical graphs.
Thus, due to the mentioned methodology's advantages and supporting robotics interface for feasible integration, it has been employed in the current paper for an end-to-end drone-based \ac{VSLAM} system.
\section{Proposed Approach}
\label{sec_proposed}

This section discusses the hardware and software components this paper utilizes for the integration procedure of a \ac{VSLAM} framework into a drone.

\subsection{Visual \ac{SLAM} Framework}
\label{sec_vslam}

The framework selected for this work \cite{vsgraphs} provides more comprehensive reconstructed maps along with their multi-level topological graphs and semantic information derived from fiducial markers.
It has been built upon ORB-SLAM3 \cite{orbslam3} and adds the required modules to detect ArUco markers, estimate the pose of the semantic objects labeled with the markers, and perform mapping and optimizing based on them.
The mentioned approach leverages the ArUco markers affixed to construction-level semantic entities, including walls and doors, reconstructs the semantic map with higher-level semantic entities, including rooms and corridors, and provides hierarchical representations.
It should be noted that mapping the high-level semantic entities requires detecting structure-level objects with the help of fiducial markers, along with a database of abstract semantic information about the topological affiliations of the walls and doors labeled with markers.

Fig.~\ref{fig_pipeline} depicts the constituent system components and the procedure flow in the mentioned \ac{VSLAM} system.
Accordingly, the framework employs a multi-thread architecture for processing data, including \textit{tracking}, \textit{local mapping}, \textit{loop and map merging}, \textit{bundle adjustment}, and \textit{semantic analysis}.
The input visual data provided by an RGB-D camera passes to the fiducial marker detection library and the \textit{tracking} module for extracting ORB features and generating KeyFrame candidates with pose information and 3D map points in the camera and global reference frames.
The \textit{local mapping} thread refines the map based on the newly added KeyFrames, markers, and 3D points and triggers the \textit{semantic analysis} module to identify the type of the detected structure-level objects and map them.
It should be mentioned that the fetched semantic information also indirectly improves the \textit{local mapping} and \textit{KeyFrame culling} modules.
Finally, as the system is constantly cooperating with \textit{Atlas} map manager, \textit{loop and map merging} is handled within active/inactive maps, and local/global \textit{bundle adjustments} is triggered regarding the optimization needed for the reconstructed map.

As the mentioned framework supports \ac{ROS}, integrating it into an aerial robot requires fine-tuning the parameters and configurations and providing proper \textit{topics} to feed camera data.
\subsection{Robot Integration and Adaptation}
\label{sec_proposed_integration}

\begin{figure}[t]
    \centering
    \includegraphics[width=0.99\columnwidth]{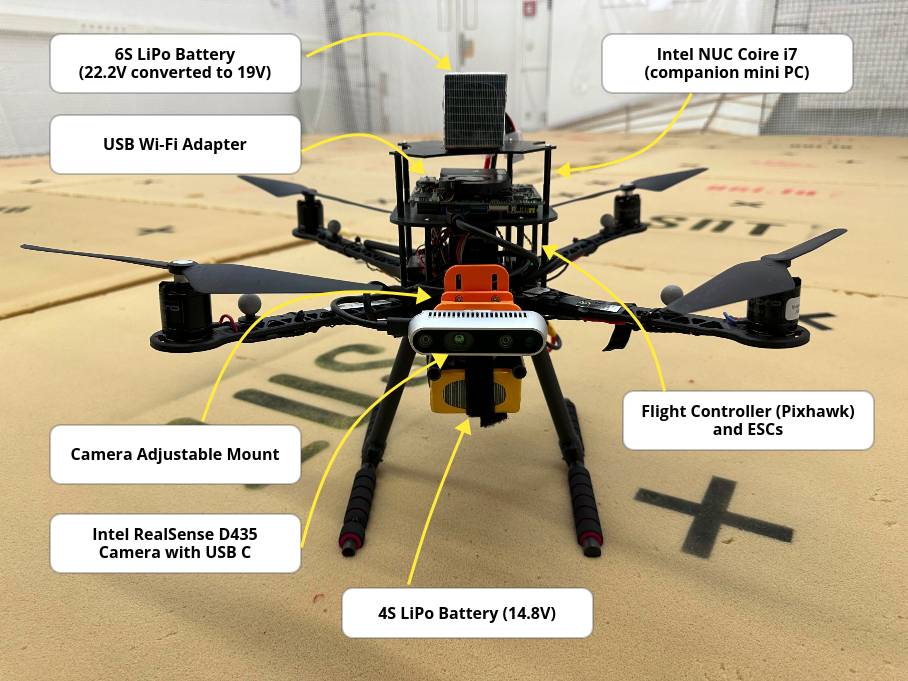}
    \caption{The drone used in the research for integration with the \ac{VSLAM} framework and experiments.}
    \label{fig_setup}
\end{figure}

The aerial robot setup provided in this work for integration with the \ac{VSLAM} framework and conducting evaluations are shown in Fig.~\ref{fig_setup}.
Accordingly, the drone is a \textit{Mikrokopter2} quadrotor built upon the Holybro S500 Quadcopter frame kit with carbon fiber landing gear.
It has four Air Gear 450 Combo 2216 Kv880 motors, a LiPo 8000 lite 4S battery pack with 118.4 Wh, and a Flysky controller offering three flight modes (\ie alt. hold, manual, and Land).
The Pixhawk 4 flight controller of the drone is used to acquire \ac{IMU} measurements at the rate of 1 kHz adapted to its motor controllers.
The vision sensor mounted on the drone is an Intel® RealSense™ Depth Camera D435, providing $87\degree\times58\degree$ Field of View (FoV) and global shutter frame acquisition.
An Intel® NUC Kit NUC5i7RYH without housing is mounted onboard as a companion computer, and a desktop computer is used as the \ac{GCS} for mission commanding and output visualization.
The drone has a payload of 1 kg, which carries the 6S battery ($\sim$590g), the mini PC ($\sim$220g), and the camera ($\sim$75g), a total of $\sim$885g of payload.
Accordingly, the takeoff of the drone requires $\sim$85\% battery throttling.

The firmware programming of the drone contained flashing PX4 and uploading to the Pixhawk, configuring the frame type, flight-related parameters fine-tuning, calibrating the sensors, and setting up the communication protocols.
The companion computer was equipped with a \ac{ROS} Noetic and the required libraries and frameworks for running the system and communicating with the \ac{GCS}.
With the mentioned setup, the \ac{VSLAM} framework can feasibly work on the drone in both live (online) and offline modes.
It should be added that the camera calibration parameters and the dimensions of the fiducial markers are fetched for accurate 3D pose estimation \wrt the camera frame.

\begin{figure}[t]
    \centering
    \includegraphics[width=0.95\columnwidth]{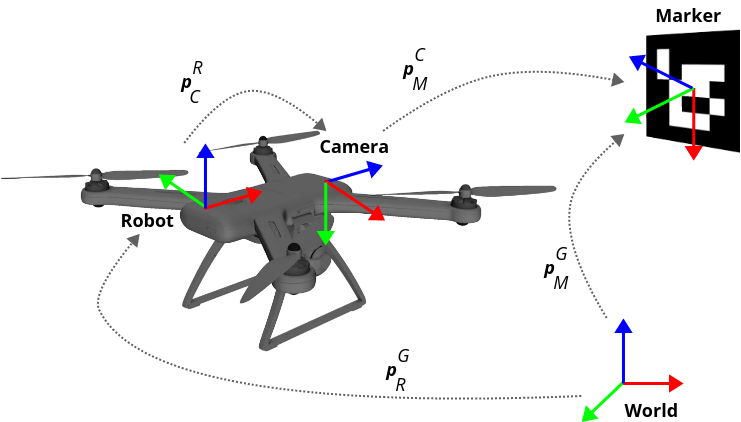}
    \caption{The reference frames considered in the current framework for perception calculations.}
    \label{fig_frames}
\end{figure}

As can be seen in Fig. \ref{fig_frames}, there are various conversions for determining the transformations in the system among the robot, camera, marker, and world reference frames.
Accordingly, robot-to-world ($p_R^G$) transformation is done using $p_R^G = T_{R}^{G} \cdot p_R$, where $T_{R}^{G}$ is the transformation matrix between the robot and the world and $p_R$ is the coordinates of a point in the world reference frame.
Similarly, camera-to-robot ($p_C^R$) transformation is calculated using $p_C^R = T_{C}^{R} \cdot p_C$ for the transformation matrix between the camera and the robot $T_{C}^{R}$ and the point in the camera reference frame $p_C$.
For marker-to-camera ($p_M^C$) and marker-to-world ($p_M^G$) transformations, $p_M^C = T_{M}^{C} \cdot p_M$ and $p_M^G = T_{M}^{G} \cdot p_M$ are calculated, where $T_{M}^{C}$ and $T_{M}^{G}$ the transformation matrix between "the marker and the camera" and "the marker and the world," and $p_M$ is the coordinates of a point in the marker reference frame.

These calculations can aid in coordinating among different reference frames using matrix multiplication and Quaternion operations.
It should be added that in the world reference frame, the center is located on the ground with a vertical $z$-axis.
In contrast, in the marker reference frame, the normal vector of the ArUco marker is parallel to the $z$-axis with a downward $x$-axis.
The robot and camera reference frames are rigidly attached to the \ac{UAV} and the mounted visual sensor with $z$-axis values perpendicular to the body and the acquired image plane, respectively.
\section{Evaluation}
\label{sec_evaluation}

This section discusses the evaluation setup and experimental results obtained by employing the designed system.
It should be emphasized that the experiments are not conducted to evaluate the performance of the employed \ac{VSLAM} methodology but instead showcase its functionality and capability in real-world indoor scenarios as an end-to-end situationally-aware \ac{UAV} system.

\subsection{Evaluation Setup}
\label{sec_eval_setup}

To evaluate the proposed \ac{UAV}-assisted \ac{VSLAM} system in real-world scenarios, the robot introduced in Section~\ref{sec_proposed_integration} was employed.
For collecting data, the walls and door frames of various indoor environments with dissimilar configurations and layouts were labeled with unique \(10cm \times 10cm\) ArUco markers.
Additionally, the topological relations among the markers were recorded and provided as a feed to the framework's \textit{Semantic Perception} database shown in Fig.~\ref{fig_pipeline}.

\begin{table}[t]
    \scriptsize
    \centering
    \caption{The characteristics of the collected indoor dataset. \textit{Set-07} to \textit{Set-09} were collected in a standard drone-testing arena with accurate ground-truth values provided.}
    \begin{tabular}{l | c | p{5.2cm}}
        \toprule
        \textbf{\textit{Data}} & \textbf{\textit{Duration}} & \textbf{\textit{Description}} \\
        \midrule
            \textit{Set-01} & 2m 02s & A single room with a door \\
            \textit{Set-02} & 3m 32s & A long corridor with parallel walls and five doors \\
            \textit{Set-03} & 2m 49s & A room within another room (nested) \\
            \textit{Set-04} & 4m 22s & A room with different nested partitions \\
            \textit{Set-05} & 1m 50s & A corridor connected to other rooms and corridors \\
            \textit{Set-06} & 1m 57s & A single room with connected doors \\
            \midrule
            \textit{Set-07} & 2m 06s & AeRoLab Drone-Testing Arena \\
            \textit{Set-08} & 1m 57s & AeRoLab Drone-Testing Arena \\
            \midrule
            \textbf{Total} & 20m 35s & \\
        \bottomrule
    \end{tabular}
    \label{tbl_dataset}
\end{table}

\begin{figure}[t]
    \centering
    \includegraphics[width=0.95\columnwidth]{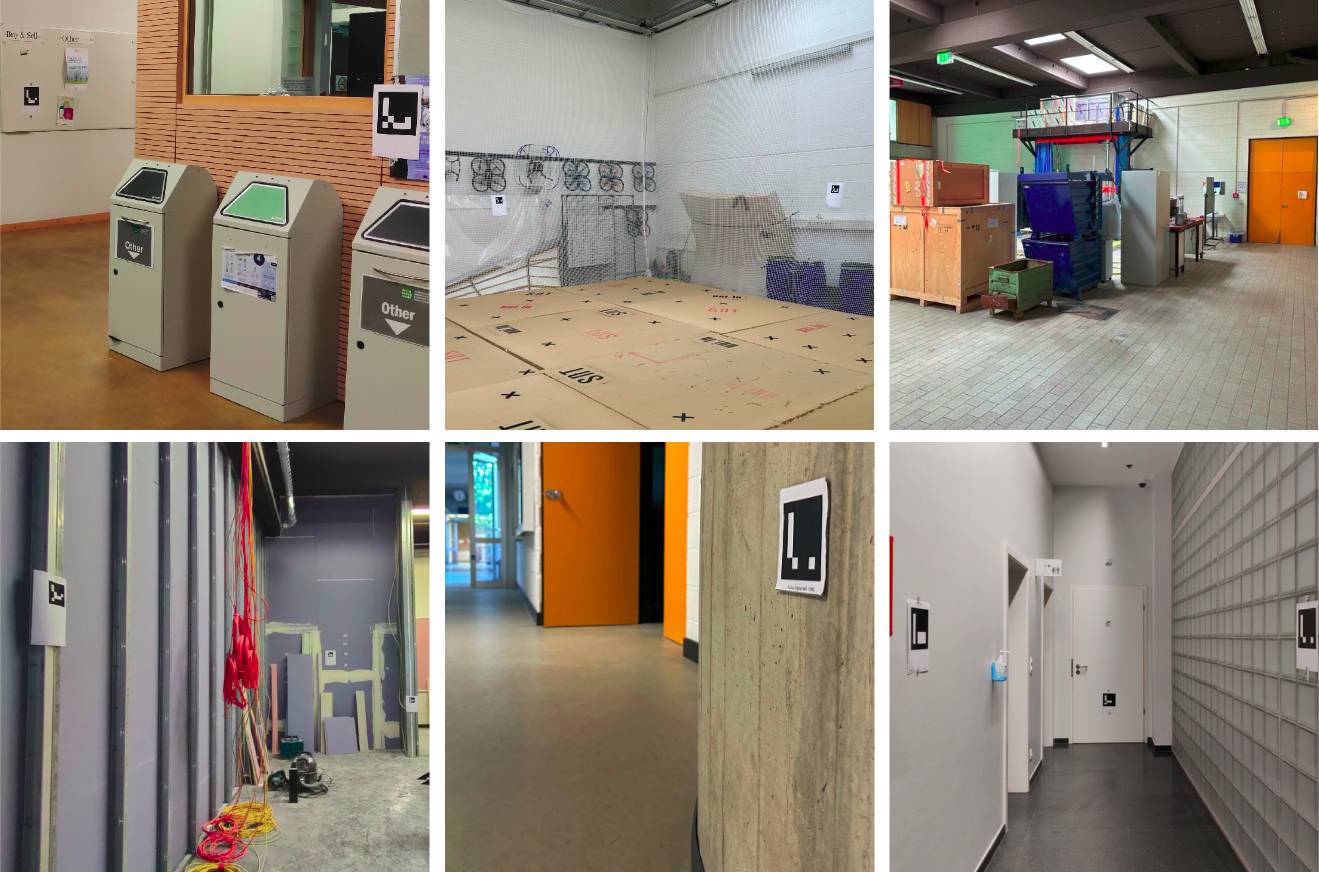}
    \caption{Some instances of the collected dataset, where ArUco markers are used to label walls and door frames of the environments.}
    \label{fig_dataset}
\end{figure}

Table~\ref{tbl_dataset} presents the collected dataset's characteristics for conducting evaluations.
According to the Table, and as shown in Fig.~\ref{fig_dataset}, various indoor area setups have been labeled with printed fiducial markers.
It should be mentioned that for experiments conducted in the drone testing arena (\ie \textit{Set-07} and \textit{Set-08}), the installed OptiTrack motion capture system in the laboratory provides ground-truth data.
The motion capture system contains twelve high-speed infrared cameras illuminating the reflective markers attached to the robot that flies within the capture volume.
Ground-truth data is provided by identifying and tracking the mounted reflective marker in sequential frames and recording the pose of the flying drone using OptiTrack.
However, as it is not possible to record ground-truth data in locations outside the drone-testing arena, other dataset instances are mainly collected to evaluate the performance of the proposed methodology \wrt other \ac{VSLAM} benchmarking frameworks in real-world scenarios.
\subsection{Scene Graphs}
\label{sec_eval_graphs}

Reconstructing the maps of the indoor environments formed by considering geometric and semantic primitives, as well as providing their 3D hierarchical representations enriched with semantic entities, is the primary demand in this paper.
This can show the practicality and functionality of the proposed robotics system, which contains a \ac{VSLAM} software integrated into a drone, and its potential in considering it as an end-to-end application for indoor environment mapping.

\begin{figure*}[t]
    \centering
    \includegraphics[width=1.0\textwidth]{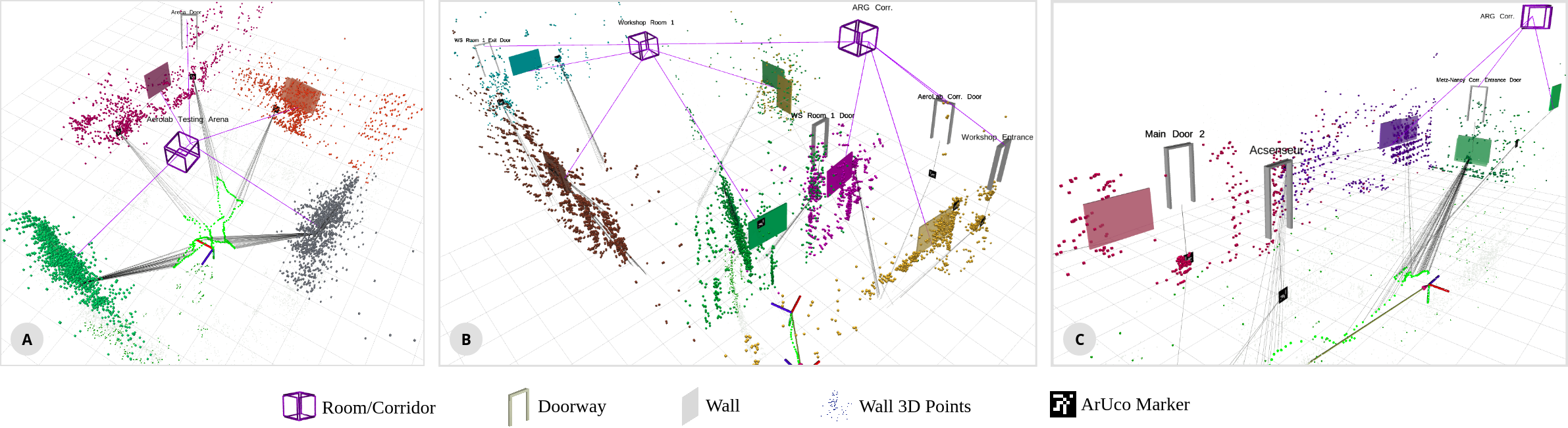}
    \caption{Some instances of the 3D scene graphs generated by the employed \ac{VSLAM} framework integrated with a \ac{UAV}: \textit{A) Set-01}, \textit{B) Set-03}, and \textit{C) Set-05}.}
    \label{fig_scene_graph}
\end{figure*}

Accordingly, Fig.~\ref{fig_scene_graph} depicts some instances of generated 3D scene graphs using the employed \ac{VSLAM} framework.
The generated hierarchies contain various abstraction levels of semantic entities detected in the environment by the employed system.
It should be noted that the correct state of the aerial robot obtained from the visual sensor is required to perform drift-free state estimation and \ac{VSLAM}.
The figure shows that the walls and door frames are detected and mapped to the global reference frame using the pose of the fiducial markers attached to them.
Colored frames and the 3D points with the same color represent the walls and their detected visual features.
Detection of rooms (with four walls) and corridors (with two parallel walls), which are shown as cubes, is the next stage handled by the \ac{VSLAM}.
It can be seen that the generated scene graphs give a high-level representation of the environment in which the drone is performing.

\subsection{Experimental Results}
\label{sec_eval_evaluation}

Various experiments have been conducted to validate the functionality of the proposed end-to-end system in \ac{GPS}-denied environments.
For evaluating the accuracy of the reconstructed maps, \ac{APE} measurements, including Mean, \ac{RMSE}, and \acf{STD} metrics, are used
Experiments are divided into two main categories described below.

\textbf{Performance \wrt the Ground-truth.}
\label{sec_eval_evaluation_gt}
In this experiment, dataset instances with provided ground-truth values obtained from the OptiTrack motion capture system are employed.
The primary goal of the experiment is to figure out how accurately the proposed \ac{VSLAM} framework performs for the correct reconstruction of the maps.
For this purpose, the data instances were fed to the employed \ac{VSLAM} framework and ORB-SLAM 3.0 as the baseline, and mean, \ac{RMSE}, and \ac{STD} values were calculated.

In this regard, Table~\ref{tbl_evaluation_gt} presents the evaluation results of flying the drone in a standard drone-testing arena where the ground-truth is available.
According to the table, the employed \ac{VSLAM} framework works better than its baseline method in both dataset instances.
The primary reason for such improvement is applying the poses obtained from fiducial markers and the structural elements (\ie door frames and walls) to optimize the map.
Although the improvement is slight, it can result in an average increase of $6.99\%$ in \ac{RMSE} and $12.21\%$ in mean measurements.
The average \ac{STD} value of the baseline is better than the employed \ac{VSLAM}, which can be due to the impact of fiducial markers in estimations.
It should be noted that these experimental results are apart from the advantage of the employed \ac{VSLAM} system in providing 3D scene graphs and hierarchical representation of the environment, which will be discussed in \textit{relative performance evaluations}.

\begin{table}[t]
    \centering
    \caption{Evaluation results on the collected data instances \wrt ground-truth data provided by the OptiTrack motion capture system. The employed metrics are Mean, \acf{RMSE} error in \textit{meters}, and \acf{STD}. The best results are boldfaced.}
    \begin{tabular}{l | ccc | ccc }
        \toprule
            & \multicolumn{3}{c}{Ours \cite{vsgraphs}} & \multicolumn{3}{c}{ORB-SLAM 3.0 \cite{orbslam3}} \\
            \cmidrule{2-7}
            & \textit{Mean} & \textit{RMSE} & \textit{STD} & \textit{Mean} & \textit{RMSE} & \textit{STD} \\
            \midrule
        \multicolumn{1}{l|}{\textit{Set-07}} & \textbf{0.2267} & \textbf{0.2689} & 0.1447 & 0.251 & 0.2757 & \textbf{0.1141} \\
        \multicolumn{1}{l|}{\textit{Set-08}} & \textbf{0.2041} & \textbf{0.2301} & \textbf{0.1027} & 0.2401 & 0.2617 & 0.1042 \\
        \midrule
        \multicolumn{1}{l|}{\textbf{\textit{Overall}}} & \textbf{0.2158} & \textbf{0.2502} & 0.1245 & 0.2458 & 0.269 & \textbf{0.1093} \\
        \midrule
        \multicolumn{1}{l|}{\textit{\textbf{Improved}}} & 12.21\% & 6.99\% & -12.21\% & - & - & - \\
        \bottomrule
    \end{tabular}
    \label{tbl_evaluation_gt}
\end{table}

\begin{figure}[t]
     \centering
     \begin{subfigure}[t]{0.95\columnwidth}
         \centering
         \includegraphics[width=\textwidth]{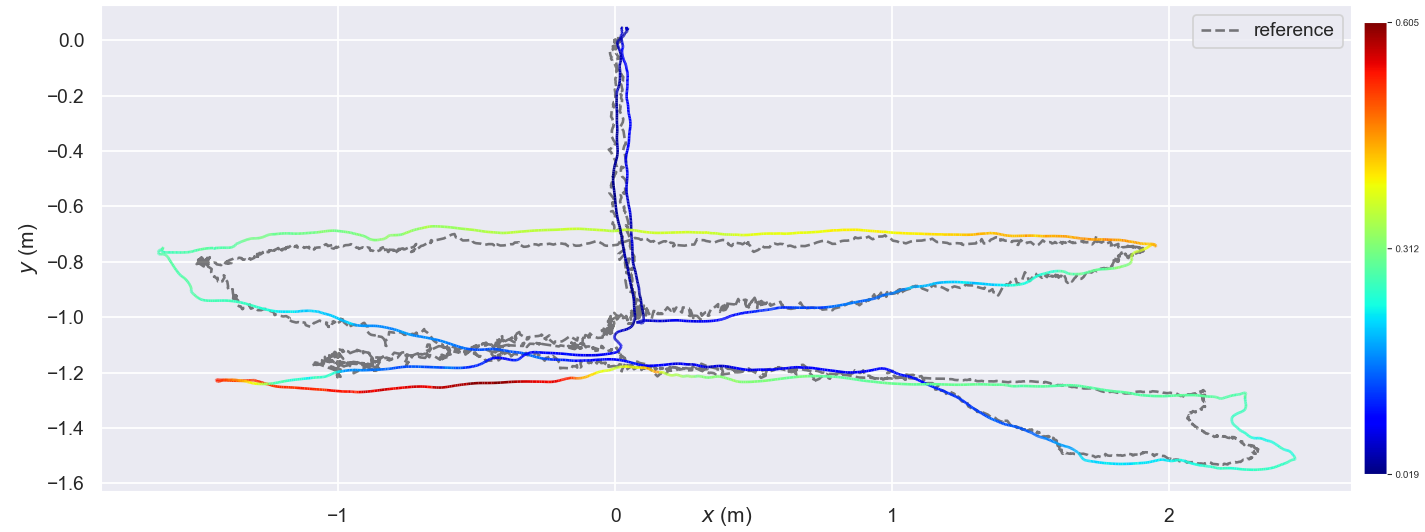}
         \caption{Evaluation using the proposed \ac{VSLAM} on \textit{Seq-07}}
         \label{fig_eval_vsg_s07}
     \end{subfigure}
     \vfill
     \begin{subfigure}[t]{0.95\columnwidth}
         \centering
         \includegraphics[width=\textwidth]{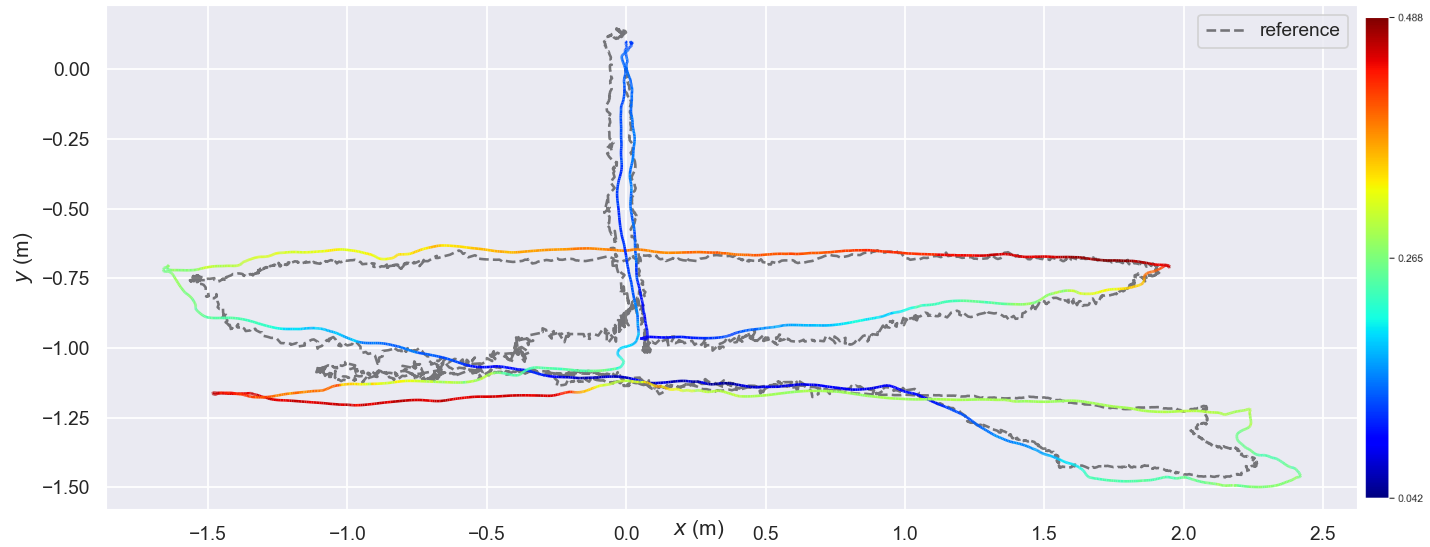}
         \caption{Evaluation using ORB-SLAM 3.0 \cite{orbslam3} on \textit{Seq-07}}
         \label{fig_eval_orb_s07}
     \end{subfigure}
     \vfill
     \begin{subfigure}[t]{0.95\columnwidth}
         \centering
         \includegraphics[width=\textwidth]{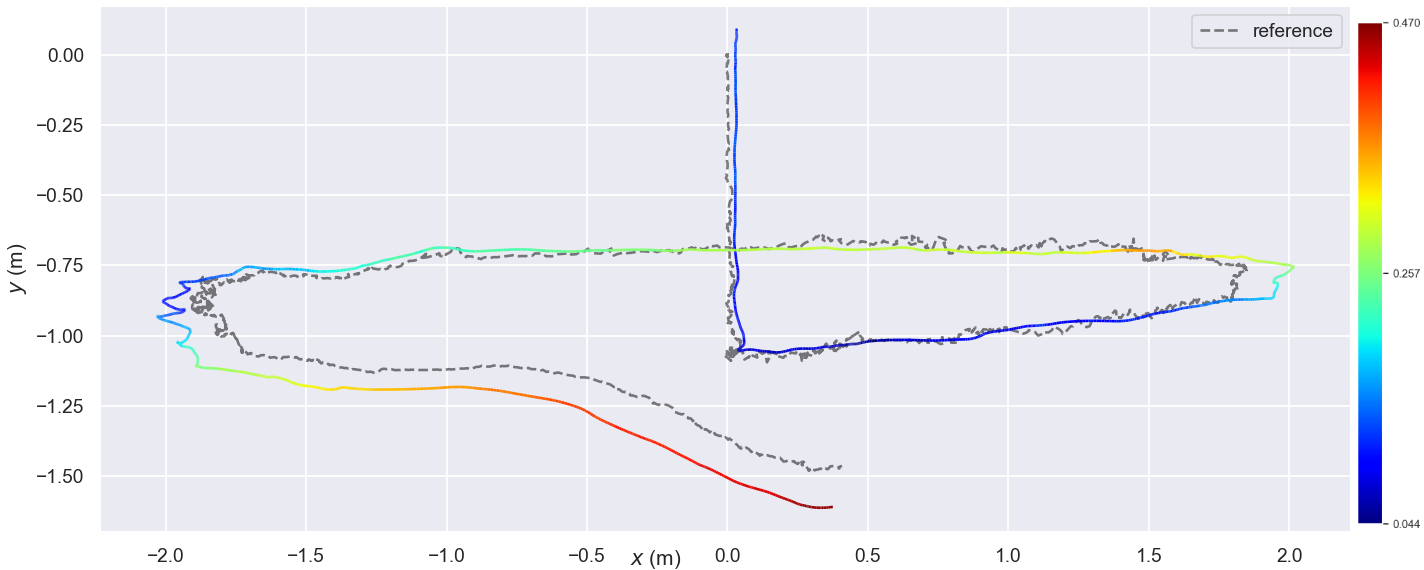}
         \caption{Evaluation using the proposed \ac{VSLAM} on \textit{Seq-08}}
         \label{fig_eval_vsg_s08}
     \end{subfigure}
     \vfill
     \begin{subfigure}[t]{0.95\columnwidth}
         \centering
         \includegraphics[width=\textwidth]{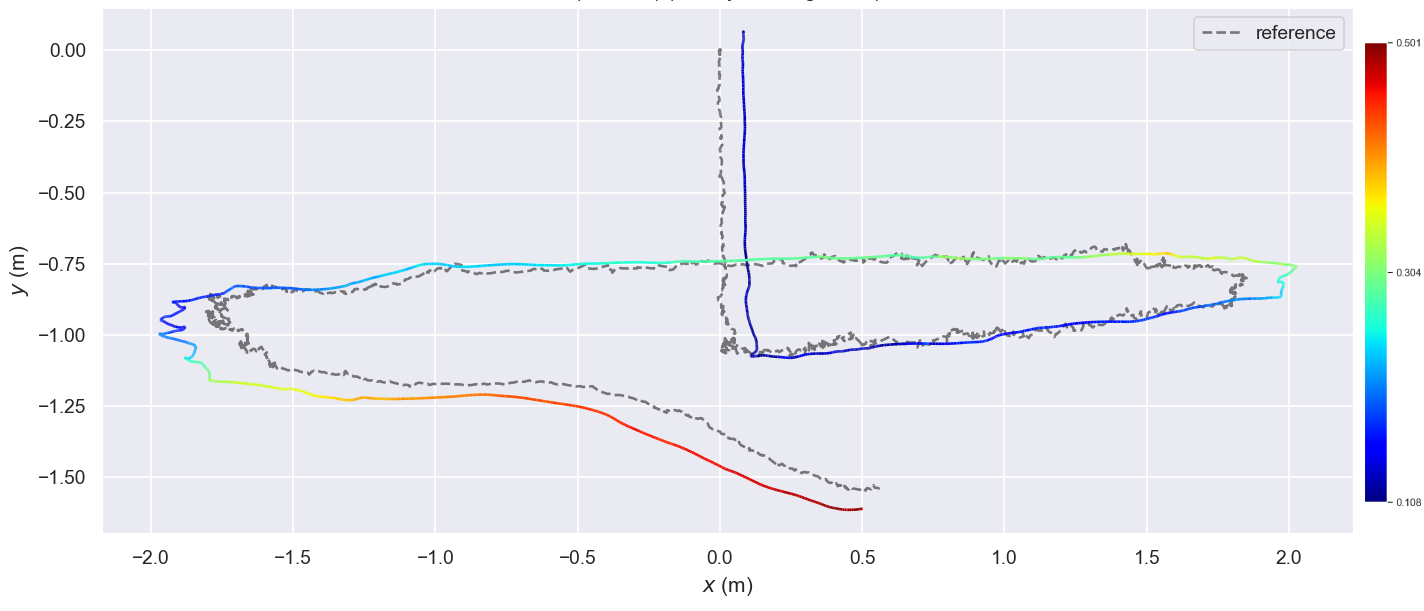}
         \caption{Evaluation using ORB-SLAM 3.0 \cite{orbslam3} on \textit{Seq-08}}
         \label{fig_eval_orb_s08}
     \end{subfigure}
     \vfill
     \caption{Trajectories traversed using the proposed system in the drone-testing arena. The dotted lines show the ground-truth values fetched by the OptiTrack motion capture system.}
     \label{fig_evals_gt}
\end{figure}

Additionally, Fig.~\ref{fig_evals_gt} depicts the trajectories generated by the drone experiments, in which the dotted trajectories are ground-truth and the colored ones are estimated.
As shown in Table~\ref{tbl_evaluation_gt}, the trajectories estimated by the employed framework seem closer to the ground-truth path in some cases.

\textbf{Relative Performance \wrt the Baseline.}
\label{sec_eval_evaluation_relate}
These experiments are mainly conducted in indoor environments where ground-truth data is unavailable, and the evaluations are obtained by comparing the employed \ac{VSLAM} methodology \wrt ORB-SLAM 3.0.
The primary targets of conducting such experiments are to provide a qualitative assessment of the system, analyze the impact of modifications and adding arbitrary features and semantic entities on the generated outputs, and seek differences in the generated maps.
Table~\ref{tbl_evaluation_rel} presents the evaluation results of employing the proposed \ac{VSLAM} system \wrt the baseline it has been built upon.

\begin{table}[t]
    \centering
    \caption{Relative performance evaluation results using the proposed \ac{VSLAM} system \wrt the ORB-SLAM 3.0 on the collected data instances. The metrics used are Maximum, Minimum, Mean, \acf{RMSE} error in \textit{meters}, \ac{SSE}, and \acf{STD}.}
    \begin{tabular}{l | cccccc }
        \toprule
            & \textit{Max.} & \textit{Mean} & \textit{Min.} & \textit{RMSE} & \textit{SSE} & \textit{STD} \\
            \midrule
        \multicolumn{1}{l|}{\textit{Set-01}} & 0.5276 & 0.0979 & 0.0363 & 0.1108 & 44.6885 & 0.0519 \\
        \multicolumn{1}{l|}{\textit{Set-02}} & 0.8105 & 0.0905 & 0.0239 & 0.1278 & 90.9611 & 0.0903 \\
        \multicolumn{1}{l|}{\textit{Set-03}} & 0.9467 & 0.3403 & 0.0404 & 0.3802 & 623.3242 & 0.1694 \\
        \multicolumn{1}{l|}{\textit{Set-04}} & 4.3524 & 0.5118 & 0.0162 & 0.9652 & 336.5995 & 0.8183 \\
        \multicolumn{1}{l|}{\textit{Set-05}} & 0.6525 & 0.0420 & 0.0028 & 0.0856 & 22.191 & 0.0746 \\
        \multicolumn{1}{l|}{\textit{Set-06}} & 0.8738 & 0.0745 & 0.0033 & 0.1261 & 52.2667 & 0.1017 \\
        \midrule
        \multicolumn{1}{l|}{\textbf{Total}} & 1.3606 & 0.1928 & 0.0205 & 0.2993 & 195.005 & 0.2177 \\
        \bottomrule
    \end{tabular}
    \label{tbl_evaluation_rel}
\end{table}

According to the table, the average \ac{RMSE} and \ac{STD} values in the comparative analysis are close.
It means that the impact of modification and adding different modules to the baseline to identify and map semantic objects still guarantees the quality of the results generated by ORB-SLAM 3.0 as a reliable baseline framework.
In other words, the results are reliable apart from making the drone more situationally aware of the environment by optimizing the generated 3D scene graphs.
In scenarios like \textit{Set-04}, where there is a nested room inside another, the \ac{RMSE} value has the highest value in the table.
This can be due to the impact of adding the poses of the structural elements to optimize the map.
The situation is reversed in less complicated scenarios, such as \textit{Set-01} and \textit{Set-05}.
\subsection{Discussions}
\label{sec_discussion}

Considering the experimental results presented above, the proposed \ac{VSLAM} framework for indoor \ac{SLAM} map reconstruction and scene graph generation can bring pros and cons.
As for the advantages, detecting fiducial markers as helpful components for scene understanding is real-time and does not impose high computational costs on the system.
This contrasts with the approaches that require deep learning-based perception and scene processing to detect semantic objects.
Moreover, all the processes related to \ac{SLAM}, structural-level object detection, and fiducial marker recognition are handled in the companion computer used in the drone.

On the other hand, depending on fiducial markers for situational awareness needs to be carefully considered.
The fiducial markers might be unreachable, damaged, or obscured in the environment, or maybe even hard to detect because of the blurry images captured due to the shakes caused by the drone while flying.
In this regard, missing a fiducial marker may lead to the misdetection of a wall or a door frame in the first step and rooms and corridors afterward.
Additionally, any noisy marker detection due to occlusion, poor lighting conditions, \, etc, can imply wrong pose information in the reconstructed maps.
As shown in Table~\ref{tbl_evaluation_gt} and Table~\ref{tbl_evaluation_rel}, the \acl{STD} value of employing the proposed \ac{VSLAM} framework can vary a lot.
High \ac{STD} values may indicate the impact of uncertainty and variability in the estimates obtained by the method, which is not desirable in \ac{SLAM} applications.
\section{Conclusions}
\label{sec_conclusions}

This paper presented an end-to-end system with an integrated \acl{VSLAM} framework into a \acf{UAV} equipped with an RGB-D visual sensor.
In this regard, the paper's primary contribution is introducing a practical solution for utilizing a drone to reconstruct the maps of indoor environments (where GPS measures are not available) alongside generating 3D scene graphs for a high-level representation of the area.
The system targets enhancing the situational awareness of the robot while reconstructing maps by incorporating fiducial markers as versatile visual features augmented into structural-semantic elements, including walls and door frames.
Detection of higher-semantic entities in the environment, including rooms and corridors, takes place with the aid of the mentioned elements.
Various real-world scenario experiments were conducted in indoor areas with dissimilar structural layouts to showcase the proposed system's practicality.
Experimental results showed the proposed application can reliably provide abstract representations of the maps \wrt the ground-truth data.

In future works, equipping the drone with the required modules to perform autonomously is the first step to consider for having an efficient end-to-end system.
Additionally, due to the detailed and information-rich maps generated by the introduced system, the authors also plan to utilize it for "localization and navigation" and "path planning" in indoor environments.
Since the topological information derived from the environment using the VSLAM system is reliable, the current system is expected to properly integrate with the mentioned applications to facilitate decision-making while maintaining the robot's situational awareness.



\bibliographystyle{IEEEtran}
\bibliography{root}

\begin{thebibliography}{10}
\providecommand{\url}[1]{#1}
\csname url@samestyle\endcsname
\providecommand{\newblock}{\relax}
\providecommand{\bibinfo}[2]{#2}
\providecommand{\BIBentrySTDinterwordspacing}{\spaceskip=0pt\relax}
\providecommand{\BIBentryALTinterwordstretchfactor}{4}
\providecommand{\BIBentryALTinterwordspacing}{\spaceskip=\fontdimen2\font plus
\BIBentryALTinterwordstretchfactor\fontdimen3\font minus \fontdimen4\font\relax}
\providecommand{\BIBforeignlanguage}[2]{{%
\expandafter\ifx\csname l@#1\endcsname\relax
\typeout{** WARNING: IEEEtran.bst: No hyphenation pattern has been}%
\typeout{** loaded for the language `#1'. Using the pattern for}%
\typeout{** the default language instead.}%
\else
\language=\csname l@#1\endcsname
\fi
#2}}
\providecommand{\BIBdecl}{\relax}
\BIBdecl

\bibitem{mishra2020drone}
B.~Mishra, D.~Garg, P.~Narang, and V.~Mishra, ``Drone-surveillance for search and rescue in natural disaster,'' \emph{Computer Communications}, vol. 156, pp. 1--10, 2020.

\bibitem{dilshad2020applications}
N.~Dilshad, J.~Hwang, J.~Song, and N.~Sung, ``Applications and challenges in video surveillance via drone: A brief survey,'' in \emph{2020 International Conference on Information and Communication Technology Convergence (ICTC)}.\hskip 1em plus 0.5em minus 0.4em\relax IEEE, 2020, pp. 728--732.

\bibitem{park2018forestry}
S.~Park, S.~Yun, H.~Kim, R.~Kwon, and J.~Ganser, ``Forestry monitoring system using lora and drone,'' in \emph{Proceedings of the 8th International Conference on Web Intelligence, Mining and Semantics}, 2018, pp. 1--8.

\bibitem{krajewski2020round}
R.~Krajewski, T.~Moers, J.~Bock, L.~Vater, and L.~Eckstein, ``The round dataset: A drone dataset of road user trajectories at roundabouts in germany,'' in \emph{2020 IEEE 23rd International Conference on Intelligent Transportation Systems (ITSC)}.\hskip 1em plus 0.5em minus 0.4em\relax IEEE, 2020, pp. 1--6.

\bibitem{vanhie2021indoor}
J.~Vanhie-Van~Gerwen, K.~Geebelen, J.~Wan, W.~Joseph, J.~Hoebeke, and E.~De~Poorter, ``Indoor drone positioning: Accuracy and cost trade-off for sensor fusion,'' \emph{IEEE Transactions on Vehicular Technology}, vol.~71, no.~1, pp. 961--974, 2021.

\bibitem{akbari2021applications}
Y.~Akbari, N.~Almaadeed, S.~Al-Maadeed, and O.~Elharrouss, ``Applications, databases and open computer vision research from drone videos and images: a survey,'' \emph{Artificial Intelligence Review}, vol.~54, pp. 3887--3938, 2021.

\bibitem{slamtosa}
H.~Bavle, J.~L. Sanchez-Lopez, C.~Cimarelli, A.~Tourani, and H.~Voos, ``From slam to situational awareness: Challenges and survey,'' \emph{Sensors}, vol.~23, no.~10, p. 4849, 2023.

\bibitem{vsgraphs}
A.~Tourani, H.~Bavle, J.~L. Sanchez-Lopez, D.~I. Avsar, R.~M. Salinas, and H.~Voos, ``Vision-based situational graphs generating optimizable 3d scene representations,'' \emph{arXiv preprint arXiv:2309.10461}, 2023.

\bibitem{3dsg}
I.~Armeni, Z.-Y. He, J.~Gwak, A.~R. Zamir, M.~Fischer, J.~Malik, and S.~Savarese, ``3d scene graph: A structure for unified semantics, 3d space, and camera,'' in \emph{Proceedings of the IEEE International Conference on Computer Vision}, 2019, pp. 5664--5673.

\bibitem{hydra}
N.~Hughes, Y.~Chang, and L.~Carlone, ``Hydra: A real-time spatial perception system for 3d scene graph construction and optimization,'' 2022.

\bibitem{sgraphs}
H.~Bavle, J.~L. Sanchez-Lopez, M.~Shaheer, J.~Civera, and H.~Voos, ``Situational graphs for robot navigation in structured indoor environments,'' \emph{IEEE Robotics and Automation Letters}, vol.~7, no.~4, pp. 9107--9114, 2022.

\bibitem{sgraphsp}
------, ``S-graphs+: Real-time localization and mapping leveraging hierarchical representations,'' \emph{IEEE Robotics and Automation Letters}, vol.~8, no.~8, pp. 4927--4934, 2023.

\bibitem{gao2021introduction}
X.~Gao and T.~Zhang, \emph{Introduction to visual SLAM: from theory to practice}.\hskip 1em plus 0.5em minus 0.4em\relax Springer Nature, 2021.

\bibitem{vpsslam}
H.~Bavle, P.~De~La~Puente, J.~P. How, and P.~Campoy, ``Vps-slam: Visual planar semantic slam for aerial robotic systems,'' \emph{IEEE Access}, vol.~8, pp. 60\,704--60\,718, 2020.

\bibitem{kalaitzakis2021fiducial}
M.~Kalaitzakis, B.~Cain, S.~Carroll, A.~Ambrosi, C.~Whitehead, and N.~Vitzilaios, ``Fiducial markers for pose estimation: Overview, applications and experimental comparison of the artag, apriltag, aruco and stag markers,'' \emph{Journal of Intelligent \& Robotic Systems}, vol. 101, pp. 1--26, 2021.

\bibitem{aruco}
S.~Garrido-Jurado, R.~Mu{\~n}oz-Salinas, F.~J. Madrid-Cuevas, and M.~J. Mar{\'\i}n-Jim{\'e}nez, ``Automatic generation and detection of highly reliable fiducial markers under occlusion,'' \emph{Pattern Recognition}, vol.~47, no.~6, pp. 2280--2292, 2014.

\bibitem{apriltag}
E.~Olson, ``Apriltag: A robust and flexible visual fiducial system,'' in \emph{2011 IEEE International Conference on Robotics and Automation}.\hskip 1em plus 0.5em minus 0.4em\relax IEEE, 2011, pp. 3400--3407.

\bibitem{ucoslam}
R.~Mu{\~n}oz-Salinas and R.~Medina-Carnicer, ``Ucoslam: Simultaneous localization and mapping by fusion of keypoints and squared planar markers,'' \emph{Pattern Recognition}, vol. 101, p. 107193, 2020.

\bibitem{sslam}
F.~J. Romero-Ramirez, R.~Mu{\~n}oz-Salinas, M.~J. Mar{\'\i}n-Jim{\'e}nez, M.~Cazorla, and R.~Medina-Carnicer, ``sslam: Speeded-up visual slam mixing artificial markers and temporary keypoints,'' \emph{Sensors}, vol.~23, no.~4, p. 2210, 2023.

\bibitem{tagSLAM}
\BIBentryALTinterwordspacing
B.~Pfrommer and K.~Daniilidis, ``Tagslam: Robust slam with fiducial markers,'' 2019. [Online]. Available: \url{https://arxiv.org/abs/1910.00679}
\BIBentrySTDinterwordspacing

\bibitem{pfin}
G.~Raja, S.~Suresh, S.~Anbalagan, A.~Ganapathisubramaniyan, and N.~Kumar, ``Pfin: An efficient particle filter-based indoor navigation framework for uavs,'' \emph{IEEE Transactions on Vehicular Technology}, vol.~70, no.~5, pp. 4984--4992, 2021.

\bibitem{budiharto2021mapping}
W.~Budiharto, E.~Irwansyah, J.~S. Suroso, A.~Chowanda, H.~Ngarianto, and A.~A.~S. Gunawan, ``Mapping and 3d modeling using quadrotor drone and gis software,'' \emph{Journal of Big Data}, vol.~8, pp. 1--12, 2021.

\bibitem{bhatnagar2020drone}
S.~Bhatnagar, L.~Gill, and B.~Ghosh, ``Drone image segmentation using machine and deep learning for mapping raised bog vegetation communities,'' \emph{Remote Sensing}, vol.~12, no.~16, p. 2602, 2020.

\bibitem{sun2023indoor}
Y.~Sun, W.~Wang, L.~Mottola, J.~Zhang, R.~Wang, and Y.~He, ``Indoor drone localization and tracking based on acoustic inertial measurement,'' \emph{IEEE Transactions on Mobile Computing}, 2023.

\bibitem{famili2022pilot}
A.~Famili, A.~Stavrou, H.~Wang, and J.-M.~J. Park, ``Pilot: High-precision indoor localization for autonomous drones,'' \emph{IEEE Transactions on Vehicular Technology}, 2022.

\bibitem{famili2020rolatin}
A.~Famili and J.-M.~J. Park, ``Rolatin: Robust localization and tracking for indoor navigation of drones,'' in \emph{2020 IEEE Wireless Communications and Networking Conference (WCNC)}.\hskip 1em plus 0.5em minus 0.4em\relax IEEE, 2020, pp. 1--6.

\bibitem{3d_scene_graph}
I.~Armeni, Z.-Y. He, J.~Gwak, A.~R. Zamir, M.~Fischer, J.~Malik, and S.~Savarese, ``3d scene graph: A structure for unified semantics, 3d space, and camera,'' in \emph{Proceedings of the IEEE International Conference on Computer Vision}, 2019, pp. 5664--5673.

\bibitem{3D_dsg}
A.~Rosinol, A.~Gupta, M.~Abate, J.~Shi, and L.~Carlone, ``3d dynamic scene graphs: Actionable spatial perception with places, objects, and humans,'' 2020.

\bibitem{scenegraphfusion}
S.-C. Wu, J.~Wald, K.~Tateno, N.~Navab, and F.~Tombari, ``Scenegraphfusion: Incremental 3d scene graph prediction from rgb-d sequences,'' 2021.

\bibitem{tourani2022visual}
A.~Tourani, H.~Bavle, J.~L. Sanchez-Lopez, and H.~Voos, ``Visual slam: What are the current trends and what to expect?'' \emph{Sensors}, vol.~22, no.~23, p. 9297, 2022.

\bibitem{orbslam}
R.~Mur-Artal, J.~M.~M. Montiel, and J.~D. Tardos, ``Orb-slam: a versatile and accurate monocular slam system,'' \emph{IEEE transactions on robotics}, vol.~31, no.~5, pp. 1147--1163, 2015.

\bibitem{orbslam2}
R.~Mur-Artal and J.~D. Tard{\'o}s, ``Orb-slam2: An open-source slam system for monocular, stereo, and rgb-d cameras,'' \emph{IEEE transactions on robotics}, vol.~33, no.~5, pp. 1255--1262, 2017.

\bibitem{orbslam3}
C.~Campos, R.~Elvira, J.~J.~G. Rodr{\'\i}guez, J.~M. Montiel, and J.~D. Tard{\'o}s, ``Orb-slam3: An accurate open-source library for visual, visual-inertial, and multimap slam,'' \emph{IEEE Transactions on Robotics}, vol.~37, no.~6, pp. 1874--1890, 2021.

\bibitem{blitzslam}
Y.~Fan, Q.~Zhang, Y.~Tang, S.~Liu, and H.~Han, ``Blitz-slam: A semantic slam in dynamic environments,'' \emph{Pattern Recognition}, vol. 121, p. 108225, 2022.

\bibitem{qian2021semantic}
Z.~Qian, K.~Patath, J.~Fu, and J.~Xiao, ``Semantic slam with autonomous object-level data association,'' in \emph{2021 IEEE International Conference on Robotics and Automation (ICRA)}.\hskip 1em plus 0.5em minus 0.4em\relax IEEE, 2021, pp. 11\,203--11\,209.

\bibitem{doherty2019probabilistic}
K.~Doherty, D.~Baxter, E.~Schneeweiss, and J.~Leonard, ``Probabilistic data association via mixture models for robust semantic slam,'' 2019.

\bibitem{sun2022multi}
Y.~Sun, J.~Hu, J.~Yun, Y.~Liu, D.~Bai, X.~Liu, G.~Zhao, G.~Jiang, J.~Kong, and B.~Chen, ``Multi-objective location and mapping based on deep learning and visual slam,'' \emph{Sensors}, vol.~22, no.~19, p. 7576, 2022.

\bibitem{guan2020real}
P.~Guan, Z.~Cao, E.~Chen, S.~Liang, M.~Tan, and J.~Yu, ``A real-time semantic visual slam approach with points and objects,'' \emph{International Journal of Advanced Robotic Systems}, vol.~17, no.~1, p. 1729881420905443, 2020.

\bibitem{yu2018ds}
C.~Yu, Z.~Liu, X.-J. Liu, F.~Xie, Y.~Yang, Q.~Wei, and Q.~Fei, ``Ds-slam: A semantic visual slam towards dynamic environments,'' in \emph{2018 IEEE/RSJ International Conference on Intelligent Robots and Systems (IROS)}.\hskip 1em plus 0.5em minus 0.4em\relax IEEE, 2018, pp. 1168--1174.

\bibitem{yang2022visual}
S.~Yang, C.~Zhao, Z.~Wu, Y.~Wang, G.~Wang, and D.~Li, ``Visual slam based on semantic segmentation and geometric constraints for dynamic indoor environments,'' \emph{IEEE Access}, vol.~10, pp. 69\,636--69\,649, 2022.

\bibitem{semuco}
A.~Tourani, H.~Bavle, J.~L. Sanchez-Lopez, R.~M. Salinas, and H.~Voos, ``Marker-based visual slam leveraging hierarchical representations,'' in \emph{2023 IEEE/RSJ International Conference on Intelligent Robots and Systems (IROS)}, 2023, pp. 3461--3467.

\end{thebibliography}

\end{document}